\documentclass[10pt,twocolumn,letterpaper]{article}

\usepackage{cvpr}              

\usepackage{comment}

\definecolor{cvprblue}{rgb}{0.21,0.49,0.74}
\usepackage[pagebackref,breaklinks,colorlinks,allcolors=cvprblue]{hyperref}

\def\paperID{} 
\def\confName{CVPR}
\def\confYear{2026}

\title{P2GS: Physical Prior-guided Gaussian Splatting for Photometrically Consistent Urban Reconstruction}

\author{Kota Shimomura$^{1}$, 
 Hidehisa Arai$^{2}$, 
Tsubasa Takahashi$^{2}$, 
Takayoshi Yamashita$^{1}$, 
Hironobu Fujiyoshi$^{1}$ \\
Chubu University$^{1}$, \quad Turing Inc.$^{2}$\\
shimo@mprg.cs.chubu.ac.jp,
hidehisa.arai@turing-motors.com\\
}

\begin{document}
\twocolumn[{
  \renewcommand\twocolumn[1][]{#1}
  \maketitle
  \begin{center}
    \includegraphics[width=\linewidth]{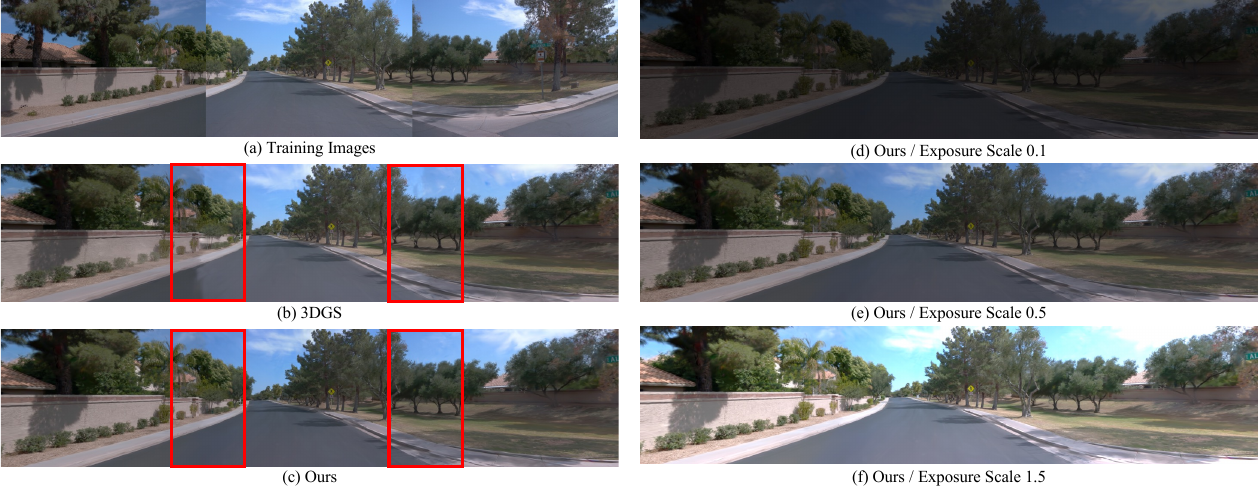}
    \captionof{figure}{We propose P2GS, a physically grounded framework that reconstructs an exposure-invariant HDR radiance field from multi-exposure driving data addressing the fundamental limitation of conventional 3DGS, which entangles radiance with camera-specific exposure and tone responses. P2GS removes exposure-induced artifacts and restores a consistent appearance across views, enabling continuous and artifact free re-rendering under arbitrary exposure scales. This exposure-aware controllability, impossible for standard 3DGS, yields photometrically stable results essential for reliable autonomous-driving simulation.}
    \label{fig:overview}
  \end{center}
}]

\begin{abstract}
3D Gaussian Splatting (3DGS) has recently emerged as a powerful explicit representation enabling fast, high-fidelity rendering, making it a promising foundation for closed-loop simulators and perception models in autonomous driving. However, conventional 3DGS implicitly assumes consistent exposure and tone mapping across views. Real driving data violates this assumption due to heterogeneous camera pipelines and dynamic outdoor illumination, baking exposure discrepancies and sensor noise into the radiance field and producing artifacts and inconsistent illumination especially in static backgrounds crucial for realistic simulation. These issues are amplified in autonomous driving, where sparse viewpoints, varying exposures, and outdoor lighting interact, while prior work mainly targets dynamic-object reconstruction and overlooks cross-view photometric consistency.
To address this limitation, we introduce \textbf{P2GS}, a physically consistent Gaussian Splatting framework that jointly decomposes a view-invariant linear HDR radiance field, per-view exposure scales, and tone-mapping functions from only LDR images without HDR supervision. P2GS employs a unified optimization strategy grounded in the physical image-formation process, enforcing relative-exposure consistency and HDR-domain radiance regularization. This yields a radiance field robust to inter-camera illumination differences while preserving the real-time efficiency of standard 3DGS. 
Experiments across real and simulated driving environments show that P2GS matches or surpasses prior methods in LDR reconstruction while providing substantially improved photometric consistency, reliable exposure normalization, and physically coherent illumination across diverse scenes.
\end{abstract}
    
\section{Introduction}
Accurate reconstruction of large-scale urban environments from multi-view imagery is essential for 3D perception, scene understanding, and autonomous driving. 
Recent progress in \textit{3D Gaussian Splatting (3DGS)}~\cite{3dgs} has made real-time, high-fidelity rendering possible through an explicit radiance and opacity representation. 
However, despite its geometric accuracy and efficiency, conventional 3DGS is fundamentally limited by its reliance on LDR pixel fitting, which entangles scene radiance with camera-specific exposure and tone responses. 
As a result, 3DGS becomes photometrically fragile to illumination changes and multi-camera inconsistencies common in autonomous driving.

Recent extensions of 3DGS have improved dynamic-scene modeling in driving environments. 
\textit{StreetGS}~\cite{streetgs} and \textit{DrivingGaussian}~\cite{drivinggs} decompose static and dynamic elements, while \textit{PVG}~\cite{pvg} introduces periodic temporal components. 
Yet none of these works address the core issue of \textbf{cross-view photometric inconsistency}. 
Real driving datasets exhibit heterogeneous ISP pipelines, per-camera exposure control, and varying outdoor illumination (\textit{cf.}~\cref{fig:overview}(a)), causing standard 3DGS to produce illumination mismatches, color shifts, and artifacts, especially in static background regions critical for simulators~\cite{hugs,mptg}.

Several studies attempt to mitigate illumination variability by augmenting 3DGS with tone mapping or exposure modeling. 
\textit{GaussHDR}~\cite{gausshdr} uses 3D/2D tone mappers but requires multi-exposure inputs and static scenes; 
\textit{SeHDR}~\cite{sehdr} synthesizes pseudo multi-exposure Gaussians but remains unstable without HDR supervision; 
\textit{Luminance-GS}~\cite{luminace} applies per-view color corrections in 2D space, lacking physical consistency. 
These approaches remain tied to the input illumination and do not generalize to large-scale, sparsely sampled driving scenes.

\vspace{-13pt}
\paragraph{Remaining Challenges.}
Overall, existing methods face three persistent limitations:
(1) dependence on controlled or dense-view environments, restricting scalability to real driving data;
(2) absence of a physically grounded radiance formulation in explicit 3D representations; and
(3) limited applicability of 2D or implicit appearance corrections to the explicit 3DGS framework. 
A unified, physically consistent approach is required to achieve robust photometric behavior in large-scale autonomous driving scenes.

\vspace{-13pt}
\paragraph{Proposal.}
We propose \textit{P2GS}, a physically grounded reformulation of Gaussian Splatting for photometrically consistent autonomous driving simulation. 
P2GS disentangles intrinsic linear HDR radiance, per-view exposure, and tone mapping directly within the 3D optimization process without HDR supervision. 
Central to the method is the \textit{Principle of Invariant Radiance (PIR)} (see~\cref{method}), which enforces consistent radiance ratios across views by estimating and constraining relative exposures in the HDR domain. 
This HDR-based formulation decouples camera-dependent photometric distortions from true scene radiance, enabling stable appearance across viewpoints and illumination conditions. 
As shown in~\cref{fig:overview}(c), P2GS produces photometrically coherent reconstructions essential for reliable large-scale simulation.
%

P2GS delivers exposure-invariant HDR radiance, enabling stable, illumination-robust 3DGS reconstruction essential for multi-camera fusion and driving simulation.

\vspace{-13pt}
\paragraph{Contributions.}
Our contributions are threefold:
\begin{itemize}
    \item We introduce \textit{P2GS}, an explicit HDR-domain extension of 3DGS that embeds the \textit{Principle of Invariant Radiance} to separate radiance from exposure and tone variations.
    \item We propose a unified optimization of HDR radiance, exposure, and tone using a photometric loss, PIR-based exposure constraints, and regularization terms—fully differentiable and compatible with standard 3DGS.
    \item P2GS demonstrates state-of-the-art photometric consistency and HDR fidelity under diverse illumination on multi-camera driving datasets with heterogeneous ISPs.
\end{itemize}

\begin{figure*}[ht]
\centering
\includegraphics[width=1.0\linewidth]{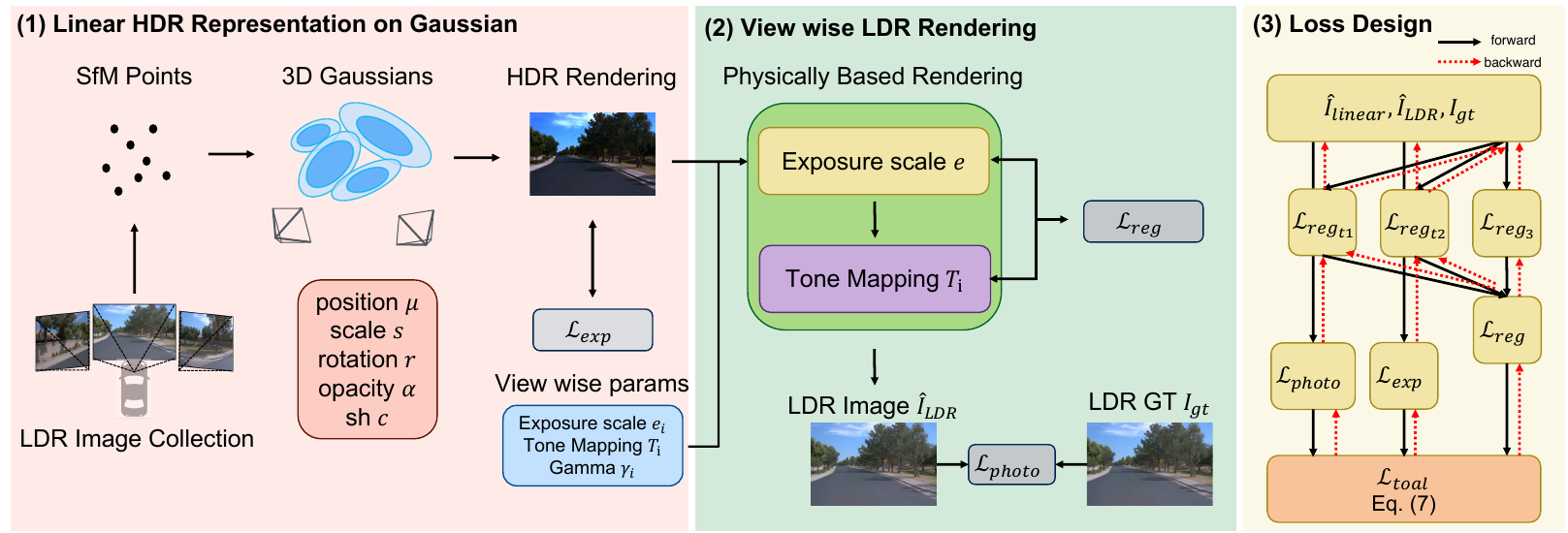}
\caption{\textbf{Overview of Our Method.} (1) To decouple the per-view exposure scales and tone-mapping functions, we learn each Gaussian as a linear HDR radiance representation. The Gaussians are constrained by a relative exposure consistency loss, which enforces the linearity of relative exposures in HDR space.
(2) LDR image rendering is then performed by applying the learned exposure and tone-mapping functions to the rendered linear HDR radiance, upon which the photometric reconstruction loss is computed.
(3) Our method is optimized by applying regularization not only to the rendered HDR and LDR images but also to the per-view physical parameters used for physically based rendering, ensuring stable and physically consistent learning across all views.}
\label{model}
\end{figure*}

\section{Related Work}
\paragraph{Neural Rendering.} 
\textit{Neural Radiance Fields (NeRFs)}~\cite{nerf} achieve high-quality 3D representations but are computationally expensive due to dense volumetric sampling and repeated MLP evaluations. 
3D Gaussian Splatting (3DGS)~\cite{3dgs} emerged as a highly efficient alternative, delivering comparable reconstruction quality while enabling real-time, tile-based differentiable rasterization.
Under varying illumination, several NeRF-based approaches model view-dependent appearance using learnable MLP embeddings~\cite{nerf1,nerf2,nerf3,nerf4,chatsim}, but such designs do not transfer cleanly to explicit representations like 3DGS. In contrast, most 3DGS methods adopt image-processing–centric brightness adjustments~\cite{sehdr,gausshdr,hdrgs}. 
However, introducing neural networks for HDR estimation or tone-curve correction undermines the core advantages of 3DGS explicitness and speed~\cite{luminace, gaussianhdr}. Furthermore, these methods typically rely on dense viewpoints and multi-exposure data, making deployment in large scale, sparse autonomous driving scenes challenging.

\vspace{-13pt}
\paragraph{3D Urban Scene Reconstruction.} 3DGS has been extended for autonomous driving to improve quality and robustness~\cite{streetgs,drivinggs,hugs,mptg,pvg}. Methods like StreetGS and DrivingGS decompose the scene into static backgrounds and dynamic vehicles using separate Gaussian sets \cite{streetgs,drivinggs}. Others achieve efficient large scale rendering and handle diverse actors by varying Gaussian properties or employing unified temporal modeling, such as PVG \cite{hugs,omnire,mptg,pvg}. PVG, which constructs a unified time-varying radiance field without static/dynamic separation \cite{pvg}, is a leading model for background representation, and is often favored in closed-loop simulators \cite{hugsim,nasim1,navsim2}.
Critically, while these methods focus on dynamics and scale, the underlying static radiance field still relies on basic 3DGS/PVG representations \cite{desiregs}. 
All these approaches implicitly assume photometric consistency, neglecting the significant issue of illumination differences between cameras in real-world driving data.

\section{Method}
\label{method}

We propose \textit{P2GS}, a physically grounded formulation of 3D Gaussian Splatting that enables
photometrically consistent reconstruction under heterogeneous exposures and tone responses.
Rather than fitting colors directly in the LDR space, P2GS shifts optimization into the \emph{linear HDR
domain}, where intrinsic scene radiance can be disentangled from camera-dependent photometric
effects. 
This HDR-domain view allows exposure, tone mapping, and radiance to be treated as
separable and jointly optimizable quantities within a coherent physical model.

To operationalize this idea, P2GS augments standard 3DGS with three components:
(i) an HDR-domain radiance representation, 
(ii) a per-view brightness module for exposure and tone modeling, and 
(iii) a unified optimization framework that enforces radiometric consistency across views.
The following sections describe these components.

\subsection{Preliminary: 3D Gaussian Splatting}\label{sec:prelim}
3DGS explicitly represents a scene as a set of $N$ learnable 3D Gaussians $\{G_k\}_{k=1}^{N}$. Each $G_k$ is characterized by a center $\boldsymbol{\mu}_k\in\mathbb{R}^3$, covariance $\boldsymbol{\Sigma}_k\in\mathbb{R}^{3\times3}$, opacity $\alpha_k\in\mathbb{R}$, and spherical-harmonics (SH) coefficients $\mathbf{c}_k$. To guarantee positive semi-definiteness, the covariance is parameterized as 
$\boldsymbol{\Sigma}_k=\mathbf{R}_k\,\mathbf{S}_k\,\mathbf{S}_k^{\top}\,\mathbf{R}_k^{\top}$,
where $\mathbf{R}_k$ is a rotation matrix and $\mathbf{S}_k$ is a scaling matrix constructed from a scale vector $\mathbf{s}_k$. This  parameterization allows independent optimization of rotation and anisotropic scaling for each Gaussian.

For rendering, Gaussians are projected onto the image plane via view and perspective transforms and then $\alpha$-blended in depth order. The color at pixel $\mathbf{p}$ is
\begin{equation}
\hat{C}(\mathbf{p})=\sum_{i\in\mathcal{N}(\mathbf{p})} c_i\,\alpha_i\,\prod_{j=1}^{i-1}\bigl(1-\alpha_j\bigr),
\end{equation}
where $\mathcal{N}(\mathbf{p})$ denotes the depth-sorted Gaussians contributing to $\mathbf{p}$, and $c_i$ is the SH-evaluated radiance of $G_i$ in the viewing direction. In conventional 3DGS, $\hat{C}$ is fitted directly to observed LDR pixel values. As a result, camera-dependent exposure and tone responses are entangled in the optimization, often breaking multi-view photometric consistency.

\subsection{Principle of Invariant Radiance (PIR)}
We introduce the \textit{Principle of Invariant Radiance (PIR)}, a physical property that enables a unified HDR-domain formulation.
PIR states that in the linear HDR domain, the ratio of observed radiances across camera views
depends solely on exposure:
\begin{equation}
\frac{I^{(i)}_{\text{linear}}}{I^{(j)}_{\text{linear}}}
= \frac{e_i}{e_j}.
\end{equation}
This invariance provides a direct constraint on HDR radiance and exposure, which we embed into optimization.

\subsection{Components of Physically Invariant Modeling}
\label{sec:components}

To achieve photometric consistency under varying illumination and camera settings, 
our method extends 3D Gaussian Splatting (Sec.~\ref{sec:prelim}) with three interdependent components: 
(1) Linear HDR-domain radiance representation, (2) exposure modeling, and (3) tone-mapping estimation. 
These components together decouple scene radiance from camera-specific photometric responses, 
laying the foundation for our unified optimization described in Sec.~\ref{sec:optimization}.

\vspace{-13pt}
\paragraph{Linear HDR-domain radiance representation.}
In conventional 3DGS, the rendered color $\hat{C}$ is fitted to observed LDR pixel intensities, 
implicitly absorbing camera exposure and tone nonlinearities into the learned SH color coefficients. 
We instead optimize each coefficient $\mathbf{c}_k$ as \textit{linear HDR radiance}, yielding a physically meaningful field $\hat{I}^i_{\text{linear}}$. 
The image formation model for view $i$ is defined as
$I_i = T_i (e_i \cdot \hat{I}^i_{\text{linear}})$,
where $e_i$ denotes the view-specific exposure scale and $T_i(\cdot)$ is the tone-mapping function. 
This formulation explicitly separates the scene-dependent radiance $\hat{I}^i_{\text{linear}}$ from per-camera photometric factors $(e_i, T_i)$, 
enabling physically interpretable optimization. 
The rendered HDR radiance at pixel $u$ is computed as
\begin{equation}
\hat{I}^i_{\text{linear}}(u) = \sum_{k\in\mathcal{K}(u)} c_k(u)\,\alpha_k'\,\prod_{j=1}^{k-1}\bigl(1-\alpha_j'\bigr),
\end{equation}
where $\mathcal{K}(u)$ denotes the Gaussians contributing to pixel $u$, $c_k(u)$ is the view-dependent SH evaluation, 
and $\alpha_k'$ is the projected opacity.

\vspace{-13pt}
\paragraph{Exposure modeling.}
Exposure is represented as a positive scalar $e_i \in \mathbb{R}^{+}$ applied to the HDR radiance:
\begin{equation}
I_{\text{exposed}}^i(u) = e_i \cdot \hat{I}^i_{\text{linear}}(u).
\end{equation}
This models the gain applied by each camera during image capture. 
The exposure parameters are initialized as $e_i\!\sim\!\mathcal{N}(1.0, 0.05^2)$ to avoid over or under-exposure at the start of training. 
Unlike previous HDR-based methods that require multi-exposure supervision~\cite{gausshdr}, 
$e_i$ is learned jointly with radiance under HDR-domain constraints, 
allowing unsupervised exposure decoupling.

\vspace{-13pt}
\paragraph{Tone-mapping estimation.}
We approximate the per-view tone-mapping response by a learnable gamma correction:
\begin{equation}
T_i(x) = \mathrm{clamp}~\!\bigl(x^{1/\gamma_i}, 0, 1\bigr),
\end{equation}
where $\gamma_i>0$ is a learnable parameter representing the nonlinear camera response. 
The derivative $\partial T/\partial \gamma_i = -(\ln x / \gamma_i^2) x^{1/\gamma_i}$ enables efficient gradient-based optimization. 
$\gamma_i$ is initialized near the sRGB prior ($\gamma \!\approx\! 2.2$) and regularized for stability. 
Before tone mapping, the exposed intensity $I_{\text{exposed}}^i$ is clamped to $[10^{-6},10]$ to prevent numerical overflow.

\vspace{-13pt}
\paragraph{Rendering for evaluation.}
At inference, $\hat{I}_{\text{linear}}$ is rendered first and then converted to an LDR image through the estimated exposure and tone parameters. 
To mitigate temporal flicker in videos or novel-view sequences, 
we average the exposure and tone parameters across all training views:
\begin{equation}
e_{\text{render}} = \frac{1}{N}\sum_{i=1}^{N} e_i, \qquad
\gamma_{\text{render}} = \frac{1}{N}\sum_{i=1}^{N} \gamma_i,
\end{equation}
which stabilizes temporal luminance variations during rendering.

\subsection{Unified HDR-Domain Optimization}
\label{sec:optimization}

We now introduce the unified optimization that jointly estimates the linear HDR radiance field, per-view exposure, and tone-mapping functions. Instead of fitting LDR colors, we optimize in the linear HDR domain under the \textit{Principle of Invariant Radiance}, ensuring that radiance ratios reflect only exposure and cleanly separating intrinsic radiance from photometric distortions without HDR supervision.

We employ a combined loss function defined as:
\begin{equation}
\mathcal{L}_{\text{total}}
= \mathcal{L}_{\text{photo}}
+ \lambda_{\text{exp}} \mathcal{L}_{\text{exp}}
+ \mathcal{L}_{\text{reg}}.
\label{eq:l_total}
\end{equation}
It comprises three complementary terms:
(1) \textit{Photometric reconstruction} $\mathcal{L}_{\text{photo}}$ — matches synthesized and observed images.
(2) \textit{Relative exposure consistency} $\mathcal{L}_{\text{exp}}$ — enforces HDR-domain exposure ratios across views.
(3) \textit{Regularization} $\mathcal{L}_{\text{reg}}$ — stabilizes exposure and tone estimation.

\vspace{-13pt}
\paragraph{Photometric reconstruction.}
Given input LDR images $\{I_i\}_{i=1}^{N}$, we minimize the photometric error between observed images and those synthesized from the estimated linear HDR radiance $\hat{I}^i_{\text{linear}}$, exposure $e_i$, and tone function $T_i$:
\begin{align}
    \mathcal{L}_{\text{photo}} = (1-\lambda_{dssim})\mathcal{L}_{1}(I_i,\hat{I}_i) + \lambda_{dssim}\mathcal{L}_{\text{DSSIM}}(I_i, \hat{I}_i).
    \label{eq:l_photo}
\end{align}

\vspace{-13pt}
\paragraph{Relative exposure consistency.}
$\mathcal{L}_{\text{exp}}$ enforces linearity of relative exposure in HDR space. 
For any view pair $(i,j)$ with relative ratio $\alpha_{ij}\!=\!e_j/e_i$, the following should hold:
$\hat{I}_{\text{linear}}^j = \alpha_{ij}\,\hat{I}_{\text{linear}}^i.$
Thus, the loss is defined as:
\begin{equation}
\begin{aligned}
\mathcal{L}_{\text{exp}} =
\frac{1}{M}\!
\sum_{(i,j)\in\mathcal{P}}
\!\left\|
\alpha_{ij}\hat{I}_{\text{linear}}^i - \hat{I}_{\text{linear}}^j
\right\|_{1},
\end{aligned}    
\end{equation}
where $\mathcal{P}$ is the set of sampled view pairs and $M$ is the number of pixels. 
This term enforces linear consistency of rendered intensities across views, 
thereby decoupling scene radiance from exposure scales. 
Unlike prior HDR-NeRF variants that handle illumination changes in the LDR space~\cite{nerf1,gausshdr}, 
this constraint preserves the physical invariance of radiance in the linear HDR domain.

\begin{table*}[t]
\centering
\caption{Our method demonstrates high reconstruction performance on the Waymo Open Dataset. FPS refers to frames per second.}
\begin{tabular}{lccccccc}
\toprule
Methods & FPS & SSIM $\uparrow$ & PSNR $\uparrow$ & LPIPS $\downarrow$ & HIS $\downarrow$ & Std-Luminance $\downarrow$ \\
\midrule
\textcolor{black!50}{Ground Truth}   & \textcolor{black!50}{-} & \textcolor{black!50}{-} & \textcolor{black!50}{-} & \textcolor{black!50}{-} & \uline{\textcolor{black!50}{\textbf{0.096}}} & \textcolor{black!50}{\textbf{0.041}} \\
\midrule
3DGS\cite{3dgs}     & 146 & \uline{0.928} & \textbf{33.62} & \uline{0.230} & 0.102 & 0.041 \\
3DGS\cite{3dgs} + Diffix\cite{diffix} & -  & 0.787 & 26.95 & 0.322 & 0.415 & 0.043 \\
PVG\cite{pvg} & 74 & 0.858 & 30.68 & 0.336 & 0.365 & 0.042 \\
PVG\cite{pvg} + Diffix\cite{diffix} & -  & 0.770 & 26.73 & 0.365 & 0.417 & 0.043 \\
\midrule
3DGS\cite{3dgs} + Affine Transform\cite{affine} & 145 & 0.927 & \uline{31.45} & 0.234 & 0.097 & \uline{0.035} \\
Luminance-GS\cite{luminace}  & - & 0.863 & 17.68 & 0.232 & 0.497 & 0.047 \\
Ours & 90 & \textbf{0.939} & 31.02 & \textbf{0.209} & \textbf{0.092} & \textbf{0.034} \\
\bottomrule
\end{tabular}
\label{table1}
\end{table*}

\vspace{-13pt}
\paragraph{Regularization.}
To prevent global-scale ambiguity in exposure, suppress variance-induced instability, and avoid non-physical tone curves, we define the sum of regularization terms as
\begin{equation}
\begin{split}
\mathcal{L}_{\text{reg}}=
\lambda_{\text{escale}}  \mathbb{E}_i[(e_i-1)^2]
+
\lambda_{\text{evar}} \mathrm{Var}(e_i) \\
+
\lambda_{\gamma} \mathbb{E}_i[(\gamma_i - \gamma_{\mathrm{prior}})^2].
\end{split}
\end{equation}
The first term resolves the global-scale ambiguity inherent in exposure estimation.
Because $\mathcal{L}_{\text{exp}}$ constrains only the relative exposure ratios
$\alpha_{ij}$, the solution is invariant under the transformation
$\{e_i\}\!\to\!\{c\,e_i\}$ for any $c>0$.
This unbounded degree of freedom allows the HDR radiance $\hat{I}_{\text{linear}}$
to be arbitrarily rescaled, undermining identifiability and destabilizing optimization.
We remove this ambiguity by softly anchoring the absolute scale of $e_i$ at $1.0$.

The second term penalizes exposure variance across views.
Large inter-view dispersion of $\{e_i\}$ can imitate tone-mapping nonlinearities
and produce brittle solutions, particularly under noisy photometric conditions or sparse view overlap.
This enforces consistent exposure behavior across cameras
while still allowing small, data-driven deviations.

The third term imposes a soft prior on per-view tone.
Unconstrained optimization of $\gamma_i$ may exploit nonlinear tone curves to absorb exposure or radiance mismatch, leading to non-physical solutions or collapse.
Centering $\gamma_i$ around the sRGB prior maintains plausible tone mapping
while retaining limited flexibility.

We set $\lambda_{\text{evar}}=\lambda_{\gamma}=0.1$ and $\lambda_{\text{escale}}=0.01$, which were chosen empirically.
Because $\lambda_{\text{evar}}>\lambda_{\text{escale}}$, the global scale is weakly anchored,
while inter-view spread is strongly regularized yielding improved stability and smoother exposure behavior.
\section{Experiments}\label{sec:experiments}

\begin{figure*}[t]
\small
\centering
\includegraphics[width=1\linewidth]{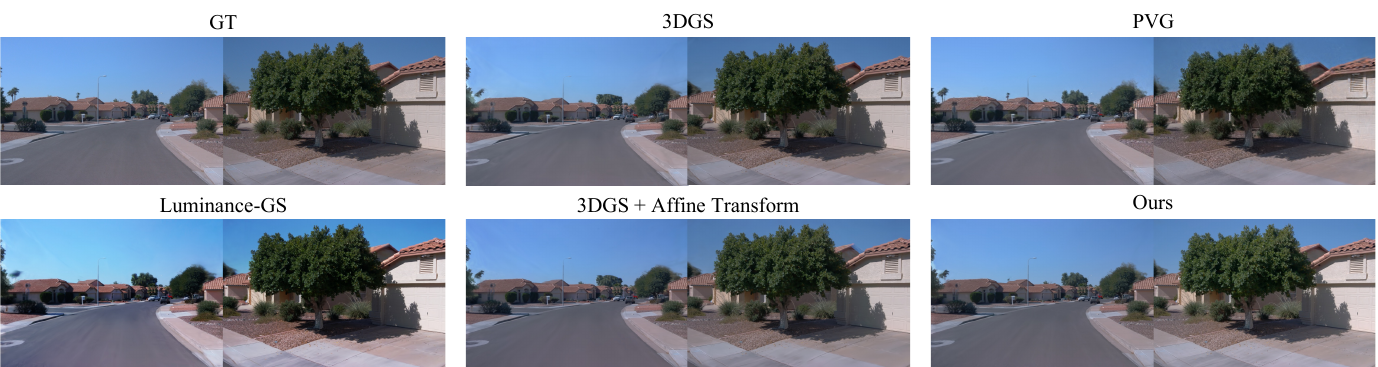}
\caption{ \textbf{Qualitative results on the Waymo Open Dataset.} Baseline methods struggle with inter-view exposure variation, producing visible seams and color inconsistencies, whereas our method maintains photometric consistency and largely suppresses seam visibility.}
\label{ex1_fig}
\end{figure*}

We evaluate \textit{P2GS} on both real-world and controlled environments. 
Since no real-world dataset provides reliable HDR ground truth, we use the Waymo Open Dataset~\cite{wod} for large-scale real driving scenes and CARLA~\cite{Dosovitskiy17} for controlled studies that isolate illumination and exposure effects. 
We assess: (1) photometric consistency and HDR fidelity of reconstructed scenes, and (2) robustness of exposure disentanglement under varying illumination.

\vspace{-13pt}
\paragraph{Competitor Methods.}~
We compare our proposed method against the vanilla 3DGS~\cite{3dgs} and PVG~\cite{pvg}, which are employed for background representation in urban scene reconstruction.
To further enhance the photometric quality of rendered images, we additionally apply Difix3D~\cite{diffix}, a post-processing method that refines the outputs of 3DGS and PVG.
For fair comparison, Difix3D is conditioned on the FRONT image of the first frame with the prompt "same illumination".
Moreover, we establish a robust baseline by adopting two independent exposure compensation methods: the state-of-the-art neural method Luminance-GS~\cite{luminace} and the affine exposure compensation method~\cite{affine}.

\vspace{-13pt}
\paragraph{Evaluation Setup.}
Following prior work~\cite{pvg}, we evaluate both reconstruction and novel view synthesis on an RTX A6000 GPU. 
For each scene, 1/8 of the frames are used for testing and the rest for training. 
Waymo scenes are initialized using SfM point clouds~\cite{Schonberger_2016_CVPR}, and PVG is modified to support this initialization. 
All baseline hyperparameters follow the original implementations.

\vspace{-13pt}
\paragraph{Training Details.}
P2GS is trained using the standard 3DGS optimization pipeline. 
We set the exposure regularization weight to
$\lambda_{\text{exp}}=0.01$, 
while all remaining training parameters follow the official 3DGS configuration. 
No HDR supervision or multi-exposure data are required.

\vspace{-13pt}
\paragraph{Metrics.}
In addition to standard image quality metrics (PSNR, SSIM, LPIPS), we introduce two measures tailored for photometric robustness in driving scenes: the \textit{HDR Inconsistency Score} (HIS) and the \textit{Standard Deviation of Luminance} (Std-Luminance). 
HIS captures temporal stability of exposure compensation, while Std-Luminance assesses inter-view brightness consistency. 
Lower values indicate more stable and reliable photometric behavior. 
Full metric definitions are provided in Appendix.

\subsection{Quantitative Results on Waymo}
\cref{table1} reports results for the image reconstruction task.
Using averaged per-view parameters for test-time rendering, our method outperforms all baselines across nearly all metrics except PSNR.
Compared with PVG, P2GS achieves notable gains of 6.9\% in SSIM, 37.7\% in LPIPS, 74.7\% in HIS, and 19.0\% in Std-Luminance.
While PSNR favors methods that reproduce the pixel-level noise present in real-world datasets such as Waymo, P2GS instead prioritizes radiometric consistency and thus delivers superior perceptual and photometric fidelity.
P2GS also surpasses the state-of-the-art neural correction method Luminance-GS~\cite{luminace} on all non-PSNR metrics.
Interestingly, it achieves HIS and Std-Luminance scores even lower than the raw ground-truth images-a consequence of the inherent inter-camera exposure inconsistency in the dataset.
This indicates that P2GS not only reconstructs high-quality images, but also effectively removes dataset-induced illumination noise that conventional 3DGS methods retain.

\cref{table2} further shows that our method maintains consistent superiority in the Novel View Synthesis (NVS) task.
By enforcing the Principle of Invariant Radiance during optimization, P2GS learns a physically coherent 3D representation that remains robust under varying lighting and exposure.
These results establish P2GS as a strong foundation for closed-loop simulation systems requiring high photometric fidelity in real-world driving environments.

\begin{table}[t]
\centering
\small
\caption{Our method shows high NVS on Waymo Open Dataset.}
\begin{tabular}{lccc}
\toprule
Methods & SSIM $\uparrow$ & PSNR $\uparrow$ & LPIPS $\downarrow$ \\
\midrule
3DGS\cite{3dgs} & \uline{0.893} & \textbf{30.39} & \uline{0.258} \\
PVG\cite{pvg} & 0.659 & 20.12 & 0.490 \\
3DGS\cite{3dgs} + AT\cite{affine} & 0.890 & \uline{28.89} & 0.263 \\
Luminance-GS\cite{luminace} & 0.630 & 16.44 & 0.427 \\
Ours & \textbf{0.896} & 28.34 & \textbf{0.246} \\
\bottomrule
\end{tabular}
\label{table2}
\end{table}

\begin{figure}[t]
\small
\centering
\includegraphics[width=0.9\linewidth]{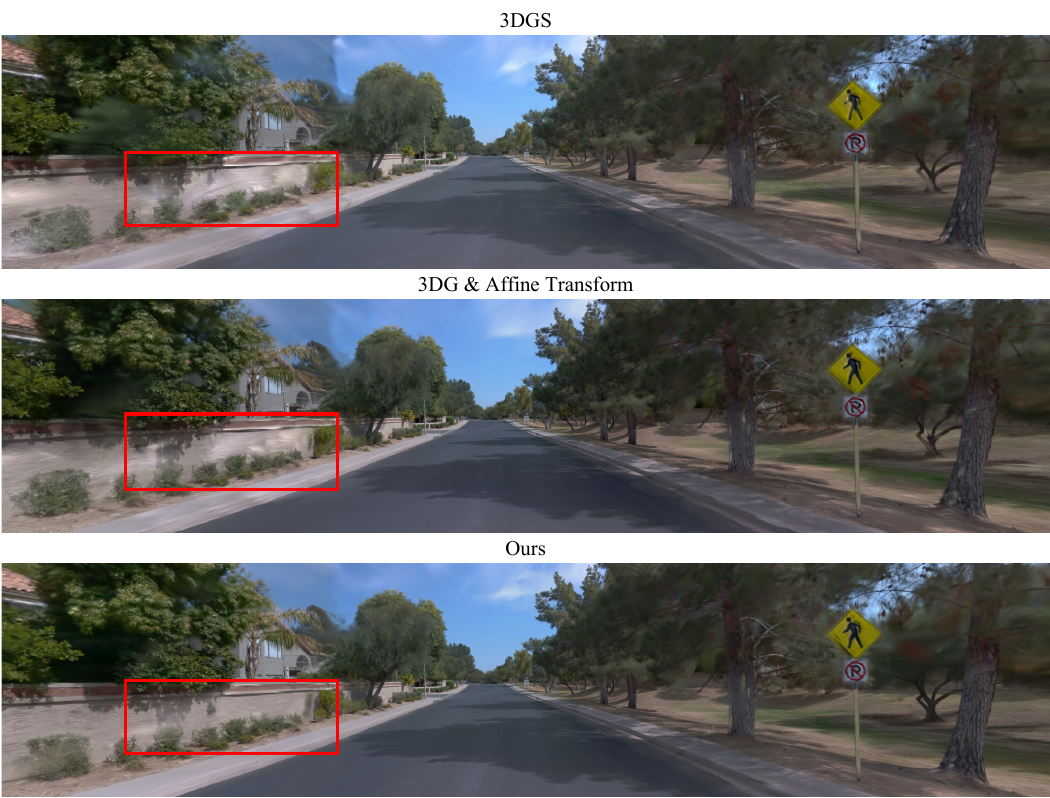}
\caption{Effects of illuminance differences between cameras. Our method yields substantially fewer artifacts under illuminance differences.}
\label{ex2_fig}
\end{figure}

\begin{table*}[t]
\small
\centering
\caption{Quantitative comparison on the CARLA Dataset. Our method preserves both robustness and reconstruction quality, even under noisy multi-camera datasets exhibiting significant inter-camera illumination discrepancies.}
\begin{tabular}{lc|cccc|cccc}
\toprule
 & & \multicolumn{4}{c|}{\textbf{Reconstruction}}  & \multicolumn{4}{c}{\textbf{Novel view synthesis}} \\
Methods & Train data 
& SSIM $\uparrow$ & PSNR $\uparrow$ & LPIPS $\downarrow$ & $\Delta\text{PSNR}$ $\uparrow$
& SSIM $\uparrow$ & PSNR $\uparrow$ & LPIPS $\downarrow$ & $\Delta\text{PSNR}$ $\uparrow$ \\
\midrule
\multirow{2}{*}{3DGS\cite{3dgs}} 
& ISO std2 & 0.842 & 21.45 & 0.273 & \multirow{2}{*}{-2.20}  & 0.814 & 19.83 & 0.290 & \multirow{2}{*}{-3.79} \\
& ISO std4 & 0.811 & 19.25 & 0.295 &  & 0.766 & 16.04 & 0.335 &  \\
\midrule
\multirow{2}{*}{PVG\cite{pvg}} 
& ISO std2 & 0.793 & 20.59 & 0.421 & \multirow{2}{*}{-1.73}  & 0.736 & 18.06 & 0.459 & \multirow{2}{*}{-0.96} \\
& ISO std4 & 0.774 & 18.86 & 0.429 &  & 0.728 & 17.10 & 0.461 &  \\
\midrule
\multirow{2}{*}{3DGS + AT\cite{affine}} 
& ISO std2 & 0.844 & 22.28 & 0.262 & \multirow{2}{*}{-3.17}  & 0.812 & 19.63 & 0.295 & \multirow{2}{*}{-3.49} \\
& ISO std4 & 0.807 & 19.11 & 0.293 &  & 0.778 & 16.14 & 0.336 &  \\
\midrule
\multirow{2}{*}{Luminance-GS\cite{luminace}} 
& ISO std2 & 0.770 & 17.36 & 0.331 & \multirow{2}{*}{-5.93}  & 0.756 & 16.99 & 0.364 & \multirow{2}{*}{-2.17} \\
& ISO std4 & 0.387 & 11.43 & 0.612 &  & 0.707 & 14.82 & 0.411 &  \\
\midrule
\multirow{2}{*}{Ours} 
& ISO std2 & \textbf{0.851} & \textbf{23.76} & \textbf{0.241} & \multirow{2}{*}{\textbf{-0.85}}  
& \textbf{0.836} & \textbf{22.72} & \textbf{0.255} & \multirow{2}{*}{\textbf{-0.95}} \\
& ISO std4 & \uline{0.847} & \uline{22.91} & \uline{0.250} & 
& \uline{0.831} & \uline{21.77} & \uline{0.262} &  \\
\bottomrule
\end{tabular}
\label{carla_ex}
\end{table*}

\begin{table}[t]
\small
\centering
\caption{Ablation Studies}
\begin{tabular}{lccc}
\toprule
 & SSIM $\uparrow$ & PSNR $\uparrow$ & LPIPS $\downarrow$ \\
\midrule
w/o $\mathcal{L}_{exp}$ & 0.920 & 27.88 & 0.237 \\
w/o $\mathcal{L}_{reg}$ & 0.920 & 27.47 & 0.234 \\
\midrule
Ours & 0.941 & 33.61 & 0.214  \\
\bottomrule
\end{tabular}
\label{ablation}
\end{table}

\subsection{Qualitative Results on Waymo} 
~\cref{ex1_fig} presents qualitative comparisons on the Waymo Open Dataset.
Baseline methods exhibit clear exposure-induced seams between views: in particular, the right half of each image often appears noticeably darker due to inter-camera exposure discrepancies. These illumination gaps propagate into the reconstructed radiance field, leading to color shifts, uneven shading, and visible patch boundaries across stitched views.

In contrast, P2GS produces photometrically uniform reconstructions with smoothly matched brightness and color across all views. The method effectively removes inter-camera exposure imbalance and suppresses seam artifacts, resulting in consistent shading on static structures such as walls, roads, and vegetation. This qualitative improvement highlights the strength of our HDR-domain formulation and demonstrates that P2GS reconstructs a physically coherent radiance field even on challenging real driving data.

~\cref{ex2_fig} further shows wide-view NVS results.
Here, we compare our method against 3DGS and the affine exposure correction method (3DGS+AT), both of which achieve relatively strong NVS accuracy.
In contrast to our approach, the conventional methods exhibit noticeable degradation in NVS quality, caused by Gaussians that encode camera-dependent illumination differences rather than scene-intrinsic radiance.

Such fidelity to inconsistent photometric conditions severely deteriorates the underlying 3D scene representation and poses a critical challenge for physically consistent closed-loop simulators.
This issue is particularly problematic when using noisy proprietary real-world datasets, yet it has been largely overlooked in prior work~\cite{desiregs,hugs}.
Our proposed method directly addresses this gap by achieving illumination-robust background reconstruction, enabling physically coherent simulation environments essential for reliable autonomous driving evaluation.

\subsection{Ablation Studies}
To verify the effectiveness of our proposed components,
we conduct ablation studies on Waymo Dataset. The
results are listed in \cref{ablation}.
Our method simultaneously decouples and optimizes the view-invariant linear HDR radiance and the view-specific exposure and tone-mapping parameters in a fully differentiable manner.
Therefore, deactivating any of the corresponding loss components results in performance degradation across all evaluation metrics.

\begin{figure*}[t]
\small
\centering
\includegraphics[width=\linewidth]{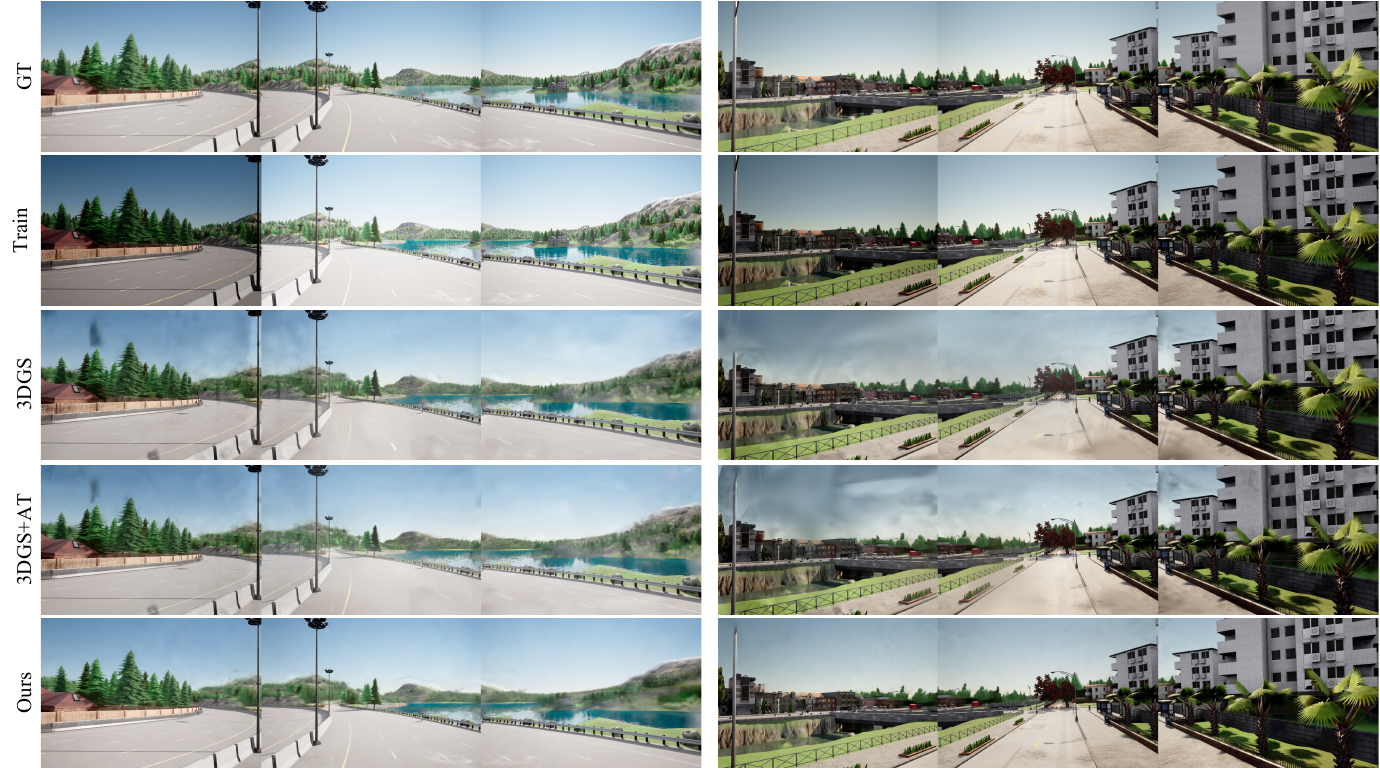}
\caption{Qualitative comparison on the CARLA dataset. 
Ground truth (GT) shows uniform illumination, while training views contain strong inter-camera exposure differences. 
Vanilla 3DGS and the affine variant (3DGS+AT) bake these inconsistencies into the Gaussians, producing noise, seams, and color shifts. 
In contrast, \textbf{P2GS reconstructs a photometrically stable radiance field consistent across views}, removing exposure-induced artifacts and restoring coherent luminance.}
\label{ex3_fig}
\end{figure*}

\subsection{Robustness against exposure noise on CARLA}
\textbf{Illumination-diverse Datasets.}
To assess radiometric consistency under exposure variation, we construct a synthetic dataset in CARLA 0.9.15~\cite{Dosovitskiy17} where exposure is the only changing factor, while geometry and camera poses remain fixed. Existing benchmarks do not revisit identical viewpoints under different exposures, making it difficult to separate photometric from geometric errors. Our dataset fills this gap by enabling evaluation of whether a method can recover a view-invariant HDR radiance field and compensate for inter-camera illumination differences while preserving LDR fidelity.
Exposure is modulated solely via ISO sensitivity, with all other camera and rendering parameters held constant. The dataset provides synchronized multi-camera sequences (front, front-left, front-right), camera intrinsics/extrinsics, and LiDAR point clouds used as 3D initialization for 3DGS.

\vspace{-13pt}
\paragraph{Experimental Setup.}
To evaluate robustness under varying illumination, we train on three synthetic datasets with distinct lighting conditions.
We evaluate the dataset captured under uniform illumination (ISO 8) as the ground truth and assess each method’s ability to restore illumination consistency from noisy training data (ISO std {2, 4}), using three distinct town scenarios in CARLA.
CARLA provides noise-free point clouds, enabling experiments under optimal data conditions.
Please refer to the Appendix for a comparison with the dynamic scene models.

\vspace{-13pt}
\paragraph{Quantitative Results.}
\cref{carla_ex} reports quantitative results on the CARLA dataset under two increasingly difficult illumination settings (ISO std2 and std4). Across both reconstruction and NVS, P2GS achieves the best overall performance and shows markedly higher robustness to illumination noise. 
When moving from ISO std2 to noisier ISO std4, P2GS exhibits only a marginal drop in SSIM and PSNR, confirming that our HDR-domain optimization effectively disentangles exposure and tone variations and preserves a view-invariant radiance field even under severe corruption.

Baseline behaviors further highlight the importance of this decoupling.
The affine transform offers slight gains under low noise but quickly fails under ISO std4, where tone and exposure distortions become nonlinear.
Luminance-GS, though effective in small or indoor scenes, does not generalize to large-scale outdoor driving with sparse viewpoints, resulting in significant degradation across metrics.

Overall, P2GS consistently delivers the highest photometric stability and reconstruction fidelity across illumination settings, establishing a physically coherent 3D representation that remains reliable even under extreme inter-camera exposure discrepancies an essential property for realistic driving simulation.

\vspace{-13pt}
\paragraph{Qualitative Results.}
\cref{ex3_fig} shows qualitative comparisons on the CARLA dataset.
Although the ground-truth frames exhibit uniform lighting, the training images contain strong inter-camera exposure differences.
Vanilla 3DGS and its affine variant (3DGS+AT) embed these discrepancies into the Gaussian primitives, producing visible seams, view-dependent color shifts, and noisy backgrounds.
Such artifacts have been largely overlooked because prior benchmarks rarely provide illumination-consistent ground truth.
In contrast, P2GS reconstructs a photometrically stable radiance field, removing exposure-induced artifacts and maintaining consistent appearance across all views.
\section{Conclusion}
We presented P2GS, an unsupervised framework that brings physical consistency to 3DGS by optimizing in the linear HDR domain. 
P2GS disentangles radiance, exposure, and tone mapping within a single explicit representation, enabling exposure-invariant reconstruction from only LDR inputs. 
By embedding the Principle of Invariant Radiance, it enforces stable relative exposure constraints and introduces HDR consistency as a new axis for evaluating Gaussian-based methods.
Although our experiments target static backgrounds, P2GS establishes a strong basis for physically coherent reconstruction in large-scale driving scenes. 
P2GS is compatible with dynamic extensions, enabling future dynamic Gaussian models.

\vspace{-13pt}
\paragraph{Limitation.}
Although P2GS reconstructs a view-invariant linear HDR radiance field, it does not explicitly separate intrinsic material properties from external illumination. 
Incorporating environmental lighting and enabling fully relightable rendering remain important future extensions.

{
    \small
    \bibliographystyle{ieeenat_fullname}
    \bibliography{main}
}

\clearpage
\setcounter{page}{1}
\maketitlesupplementary

\appendix
\section{CARLA Dataset}

We build a synthetic dataset in CARLA~0.9.15~\cite{Dosovitskiy17} to isolate exposure variation while fixing geometry, camera poses, and rendering pipeline. 
This enables controlled evaluation of view-invariant HDR radiance recovery and cross-camera photometric consistency.

\subsection{Generation Protocol}

All scenarios are generated deterministically in synchronous mode (fixed tick), with dynamic actors disabled to keep the scene static.
A single ego vehicle follows a scripted route in Town01, Town03, and Town04.
Weather, time-of-day, and illumination are fixed (e.g., clear sky and constant sun altitude). 
Sensors are time-synchronized at a constant frame rate (10\,Hz), and each sequence is 10\,s (100 frames).

\subsection{Sensor Mounting and Imaging Parameters}
\label{sec:carla_mount}

\paragraph{Camera rig.}
A rigid tri-camera rig is attached to the ego vehicle (CARLA coordinates: $x$ forward, $y$ right, $z$ up).
Per-sequence extrinsics are constant and shared across exposure settings.
Table~\ref{tab:cam_mount} lists mount poses (vehicle$\to$camera) as $(\text{loc}[m],\,\text{rot}[^\circ])$.

\begin{table}[h]
\centering
\scriptsize
\setlength{\tabcolsep}{3pt} 
\renewcommand{\arraystretch}{1.0}
\caption{Camera mounting (vehicle $\rightarrow$ camera). Location [m], rotation [deg].}
\label{tab:cam_mount}
\begin{tabular}{lcc}
\toprule
Camera & Loc $(x,y,z)$ & Rot $(\mathrm{roll},\mathrm{pitch},\mathrm{yaw})$ \\
\midrule
Front (ID 0)       & $(1.539,\,0.025,\,3.845)$ & $(0.696,\,0.420,\,0.338)$ \\
Front-left (ID 1)  & $(1.494,-0.091,\,3.845)$ & $(0.003,\,1.387,-44.205)$ \\
Front-right (ID 2) & $(1.489,\,0.095,\,3.846)$ & $(0.189,\,0.111,\,44.756)$ \\
\bottomrule
\end{tabular}
\end{table}

\paragraph{Image formation.}
Each camera records $1920\times 1300$ sRGB images at $10$\,Hz with $60^\circ$ horizontal FOV.
Intrinsics are computed from FOV and image size:
\[
f_x=f_y=\frac{W}{2\tan(\text{FOV}/2)},\quad
c_x=\frac{W-1}{2},\; c_y=\frac{H-1}{2},
\]
which yields $f_x=f_y\approx 1662.77$, $c_x=959.5$, $c_y=649.5$ for $W{=}1920$, $H{=}1300$, $\text{FOV}{=}60^\circ$.
We export KITTI-style $3{\times}4$ projection matrices $\mathbf{P}_k=\begin{bmatrix}f_x&0&c_x&0\\0&f_y&c_y&0\\0&0&1&0\end{bmatrix}$ for $k\in\{0,1,2\}$.

\paragraph{LiDAR.}
A roof-mounted LiDAR (vehicle$\to$LiDAR location $(0.0,\,0.0,\,1.73)$\,m) runs at $10$\,Hz with \emph{64} channels, $1.3$\,M points/s, range $100$\,m, vertical FOV $[-25^\circ,\,15^\circ]$, and no synthetic noise.
LiDAR scans are used only for 3D initialization.
For interoperability with KITTI-style tooling, we provide per-frame LiDAR poses and camera projection files; basis conversion to the KITTI conventions (e.g., $y$-axis flip for LiDAR) is applied when exporting calibration.

\paragraph{Photometric masks.}
For sky-sensitive analyses, we provide per-frame sky masks derived from semantic segmentation. These masks are binary and aligned to each RGB frame.

\subsection{Exposure Control and Photometric Settings}
Exposure is controlled solely via ISO under manual exposure mode. Shutter speed and aperture remain fixed, and the camera response / ISP is held constant.
We provide two subsets:
\begin{itemize}\itemsep 2pt
    \item \textbf{ISO-Const}: a fixed ISO of \textbf{8} for all frames and cameras . Illumination is constant and identical across cameras.
    \item \textbf{ISO-Var}(\textit{std}): per-frame ISO is drawn i.i.d.\ from $\mathcal{N}(\mu{=}8,\ \sigma\in\{2,4\})$, integer-rounded and lower-bounded, with the same geometry/poses as ISO-Const. This yields controlled photometric drift while preserving geometry and poses.
\end{itemize}
RGB images are recorded in sRGB (8-bit); internal linear-radiance values are used only for sanity checks.
No denoising or sharpening is applied.

\section{Details of the Optimization}
\label{sec:opt}
The proposed model is trained under a unified loss that jointly optimizes geometric, radiative, and photometric parameters. 
Since the entire pipeline is differentiable, the SH coefficients $\mathbf{c}_k$, exposure scales $e_i$, and tone-mapping gammas $\gamma_i$ are jointly optimized via automatic differentiation, enabling unsupervised separation of scene-specific and camera-specific factors.

The total loss is defined as
\begin{align}
\mathcal{L}_{\text{total}} =
\mathcal{L}_{\text{photo}} +
\lambda_{\text{exp}}\mathcal{L}_{\text{exp}} +
\mathcal{L}_{\text{reg}},
\end{align}
where $\lambda_{\text{exp}}\!=\!0.1$, and $\mathcal{L}_{\text{reg}}$ is the sum of regularization terms.

\textbf{Photometric reconstruction loss.}
$\mathcal{L}_{\text{photo}}$ minimizes the error between the observed LDR image $I_i$ and the reconstructed $\hat{I}_{\text{LDR}}^i = T_i(e_i \cdot \hat{I}_{\text{linear}}^i)$:
\begin{align}
\mathcal{L}_{\text{photo}} = (1 - \lambda_{\text{dssim}})\mathcal{L}_{L1} + \lambda_{\text{dssim}}\mathcal{L}_{\text{SSIM}},
\end{align}
where $\mathcal{L}_{L1} = \frac{1}{HW}\sum_{u}|\hat{I}_{\text{LDR}}^i(u) - I_i(u)|$ 
and $\mathcal{L}_{\text{SSIM}} = 1 - \text{SSIM}(\hat{I}_{\text{LDR}}^i, I_i)$. 
We set $\lambda_{\text{dssim}}\!=\!0.2$. 
L1 promotes pixel-level accuracy, while SSIM encourages local structural consistency, allowing optimization of both luminance and perceptual fidelity.

\textbf{Relative exposure consistency loss.}
$\mathcal{L}_{\text{exp}}$ enforces linearity of relative exposure in HDR space. 
For any view pair $(i,j)$ with relative ratio $\alpha_{ij}\!=\!e_j/e_i$, the following should hold:
\begin{align}
\hat{I}_{\text{linear}}^j = \alpha_{ij}\,\hat{I}_{\text{linear}}^i.
\end{align}
Thus, the loss is defined as
\begin{align}
\mathcal{L}_{\text{exp}} =
\frac{1}{M}\!
\sum_{(i,j)\in\mathcal{P}}
\!\left\|
\alpha_{ij}\hat{I}_{\text{linear}}^i - \hat{I}_{\text{linear}}^j
\right\|_{1},
\end{align}
where $\mathcal{P}$ is the set of sampled view pairs and $M$ is the number of pixels. 
This term requires no ground-truth exposure and directly enforces HDR-space consistency, decoupling exposure from tone nonlinearity.

\textbf{Regularization.}
To prevent global-scale ambiguity in exposure, suppress variance-induced instability, and avoid non-physical tone curves, we define
\begin{equation}
\begin{split}
\mathcal{L}_{\text{reg}}=
\lambda_{\text{escale}}  \mathbb{E}_i[(e_i-1)^2]
+
\lambda_{\text{evar}} \mathrm{Var}(e_i) \\
+
\lambda_{\gamma} \mathbb{E}_i[(\gamma_i - \gamma_{\mathrm{prior}})^2].
\end{split}
\end{equation}
Each term targets a distinct failure mode of unsupervised exposure--tone disentanglement and is weighted independently.

\textbf{\textit{(a) The first term : Global scale normalization}.}
Because $\mathcal{L}_{\text{exp}}$ constrains only relative exposure ratios $\alpha_{ij}=e_j/e_i$, the solution is invariant under the transformation $\{e_i\}\!\to\!\{c\,e_i\}$ for any constant $c>0$. 
This leaves a degree of freedom that can arbitrarily rescale the HDR radiance $\hat{I}_{\text{linear}}$, impeding identifiability and destabilizing optimization.
We remove this ambiguity by softly anchoring the absolute scale of $e_i$ at $1.0$.
The penalty centers the exposure distribution without overconstraining inter-view differences.
In practice, this term improves gradient conditioning by preventing the HDR field from absorbing arbitrary global gains.

\textbf{\textit{(b) The second term : Relative consistency via variance suppression}.}
Even with the global mean fixed, large dispersion of $\{e_i\}$ across views can mimic tone nonlinearity and cause brittle solutions, especially under limited view overlap or photometric noise.
We therefore penalize the empirical variance of exposure:
This term encourages tight clustering of exposures around a common scale while allowing small, data-driven deviations to remain.
Empirically, setting $\lambda_{\text{evar}}>\lambda_{\text{escale}}$ implements a hierarchical constraint: the global level is weakly anchored by \textit{\textbf{The first term}}, whereas \textit{\textbf{The second term}} strongly regularizes inter-view spread, yielding improved numerical stability and temporal smoothness.

\textbf{\textit{(c) The third term : Tone regularization with physical prior}.}
Unconstrained optimization of per-view gamma can exploit tone nonlinearity to explain exposure or radiance mismatch, leading to non-physical tone curves and optimization collapse.
We impose a soft prior around the sRGB standard 
The quadratic penalty preserves gradient flow and allows data-driven deviations when statistically justified, while discouraging extreme, non-physical solutions.
This term directly stabilizes the interaction between exposure scaling and tone mapping, which is critical for preserving HDR linearity in the radiance field.
The differentiable pipeline enables fully unsupervised separation of scene-specific radiance $\hat{I}_{\text{linear}}$ and camera-specific parameters $(e_i,\gamma_i)$, effectively removing camera-dependent illumination artifacts inherent in conventional 3DGS.

\section{Metrics} \label{sec:metric_detail}
In addition to standard image quality metrics (PSNR, SSIM, and LPIPS), we introduce two novel metrics designed to evaluate photometric stability and exposure robustness in 3DGS for autonomous driving scenes: 
the HDR Inconsistency Score (HIS) and the Standard Deviation of Luminance
 (Std-Luminance). 
\subsection{HDR Inconsistency Score}
HIS measures the temporal stability of exposure compensation in HDR-enabled reconstructions. 
It quantifies frame-to-frame luminance variation after exposure correction as follows:
\begin{equation}
\text{HIS} = \frac{1}{T-1} \sum_{t=1}^{T-1} \mathcal{D}_{\text{HDR}}(R_t, R_{t+1}, e_t, e_{t+1}),
\end{equation}
\begin{equation*}
\mathcal{D}_{\text{HDR}} = \left\| \text{OETF}^{-1}(R_t) \cdot e_t - \text{OETF}^{-1}(R_{t+1}) \cdot e_{t+1} \right\|_2,
\end{equation*}
where $T$ denotes the total number of frames, and $t$ indexes time. 
$R_t \in [0,1]^{H \times W \times 3}$ is the rendered image at time $t$, 
$e_t \in \mathbb{R}^+$ is the exposure scale at time $t$, and 
$\text{OETF}^{-1}: [0,1] \rightarrow \mathbb{R}^+$ denotes the inverse Opto-Electronic Transfer Function that maps sRGB to linear RGB space. 
The operator $\|\cdot\|_2$ represents the L2 norm computed over all pixels. 
Lower HIS values indicate more stable exposure compensation and higher temporal illumination consistency across frames. \\
\noindent

\begin{table*}[t]
\small
\centering
\caption{Quantitative comparison of ours versus dynamic scene models on the CARLA Dataset. Our method preserves both robustness and reconstruction and NVS quality.}
\begin{tabular}{lc|cccc|cccc}
\toprule
 & & \multicolumn{4}{c|}{\textbf{Reconstruction}}  & \multicolumn{4}{c}{\textbf{Novel view synthesis}} \\
Methods & Train data 
& SSIM $\uparrow$ & PSNR $\uparrow$ & LPIPS $\downarrow$ & $\Delta\text{PSNR}$ $\uparrow$
& SSIM $\uparrow$ & PSNR $\uparrow$ & LPIPS $\downarrow$ & $\Delta\text{PSNR}$ $\uparrow$ \\
\midrule
\multirow{2}{*}{StreetGS\cite{streetgs}} 
& ISO std2 & 0.733 & 20.16 & 0.532 & \multirow{2}{*}{-2.07}  & 0.512 & 11.04 & 0.804 & \multirow{2}{*}{-0.44} \\
& ISO std4 & 0.699 & 18.09 & 0.564 &  & 0.496 & 10.60 & 0.818 &  \\
\midrule
\multirow{2}{*}{DeSIRE-GS\cite{desiregs}} 
& ISO std2 & 0.726 & 16.94 & 0.379 & \multirow{2}{*}{-4.69}  & 0.699 & 16.35 & 0.401 & \multirow{2}{*}{-3.83} \\
& ISO std4 & 0.647 & 12.25 & 0.484 &  & 0.633 & 12.52 & 0.485 &  \\
\midrule
\multirow{2}{*}{OmniRe\cite{omnire}} 
& ISO std2 & 0.733 & 19.84 & 0.534 & \multirow{2}{*}{-1.73}  & 0.511 & 11.04 & 0.806 & \multirow{2}{*}{\textbf{-0.41}} \\
& ISO std4 & 0.699 & 18.11 & 0.565 &  & 0.498 & 10.63 & 0.823 &  \\
\midrule
\multirow{2}{*}{Ours} 
& ISO std2 & \textbf{0.851} & \textbf{23.76} & \textbf{0.241} & \multirow{2}{*}{\textbf{-0.85}}  
& \textbf{0.836} & \textbf{22.72} & \textbf{0.255} & \multirow{2}{*}{-0.95} \\
& ISO std4 & \uline{0.847} & \uline{22.91} & \uline{0.250} & 
& \uline{0.831} & \uline{21.77} & \uline{0.262} &  \\
\bottomrule
\end{tabular}
\label{ex_appen}
\end{table*}

\subsection{Standard Deviation of Luminance}
Std-Luminance evaluates global brightness consistency across all rendered views after exposure compensation. 
For each view $i$, pixel luminance $L_i(x,y)$ is computed using the ITU-R BT.601 definition:
\begin{equation}
\begin{split}
L_i(x,y) = 0.299 \cdot R_i^{(r)}(x,y) + 0.587 \cdot R_i^{(g)}(x,y) \\
+ 0.114 \cdot R_i^{(b)}(x,y),
\end{split}
\end{equation}
where $(x, y) \in \{1, \ldots, W\} \times \{1, \ldots, H\}$ are pixel coordinates, and 
$R_i^{(r)}(x,y)$, $R_i^{(g)}(x,y)$, $R_i^{(b)}(x,y) \in [0,1]$ denote the normalized RGB components of the rendered image $R_i$. 
The per-view mean luminance $\bar{L}_i$ is then computed as:
\begin{equation}
\bar{L}_i = \frac{1}{H \cdot W} \sum_{x=1}^W \sum_{y=1}^H L_i(x,y),
\end{equation}
where $\bar{L}_i \in [0,1]$ represents the average scene brightness for view $i$. 
Finally, the global brightness consistency across all $N$ views is quantified as:
\begin{equation}
\text{Std-Luminance} = \sqrt{\frac{1}{N} \sum_{i=1}^N \left(\bar{L}_i - \frac{1}{N}\sum_{j=1}^N \bar{L}_j\right)^2},
\end{equation}
where $\frac{1}{N}\sum_{j=1}^N \bar{L}_j$ denotes the global mean luminance across all views. 
Lower Std-Luminance values indicate higher inter-view brightness consistency and more reliable exposure normalization.

\section{Comparison with Dynamic Scene Models}
We compare our approach with dynamic-scene models, including StreetGS\cite{streetgs}, DeSIRE-GS\cite{desiregs}, and OmniRe\cite{omnire}. 
All baselines are trained using their publicly released implementations with hyperparameters strictly following the settings reported in their respective papers. 
Quantitative results are presented in \ref{ex_appen}. 
Across both training conditions (std2 and std4), our method consistently outperforms previous work.

In reconstruction, our approach achieves the highest $\Delta\text{PSNR}$ among all methods, demonstrating robustness even compared to dynamic-scene models that explicitly track moving objects. 
For novel-view synthesis (NVS), P2GS achieves a lower $\Delta\text{PSNR}$ than StreetGS and OmniRe.
However, it is important to note that both methods exhibit noticeably degraded NVS quality. 
In the more challenging std4 setting, our method yields substantial improvements over OmniRe, achieving gains of 62.6\% in SSIM, 97.2\% in PSNR, and 67.5\% in LPIPS.

These results highlight the strong robustness of P2GS under heterogeneous illumination, even when competing methods fail to maintain photometric consistency.

\begin{figure}[t]
\small
\centering
\includegraphics[width=\linewidth]{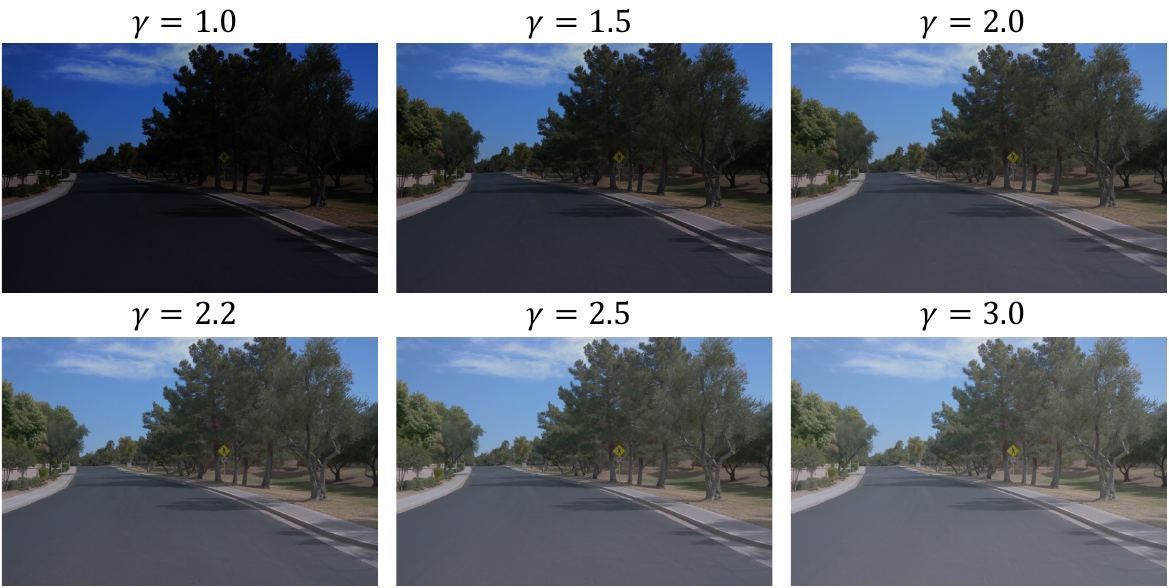}
\caption{Gamma comparison on the Waymo Open Dataset. }
\label{ap_1}
\end{figure}
\section{Application}
Our decoupling of exposure scale and tone-mapping enables rendering with arbitrary combinations of $e$ and $\gamma$. 
The key advantage is that global illumination consistency is preserved while allowing the brightness of the entire scene to be adjusted. 
This capability is independent of camera-specific characteristics or the time of data acquisition. 
\begin{figure*}[t]
\small
\centering
\includegraphics[width=\linewidth]{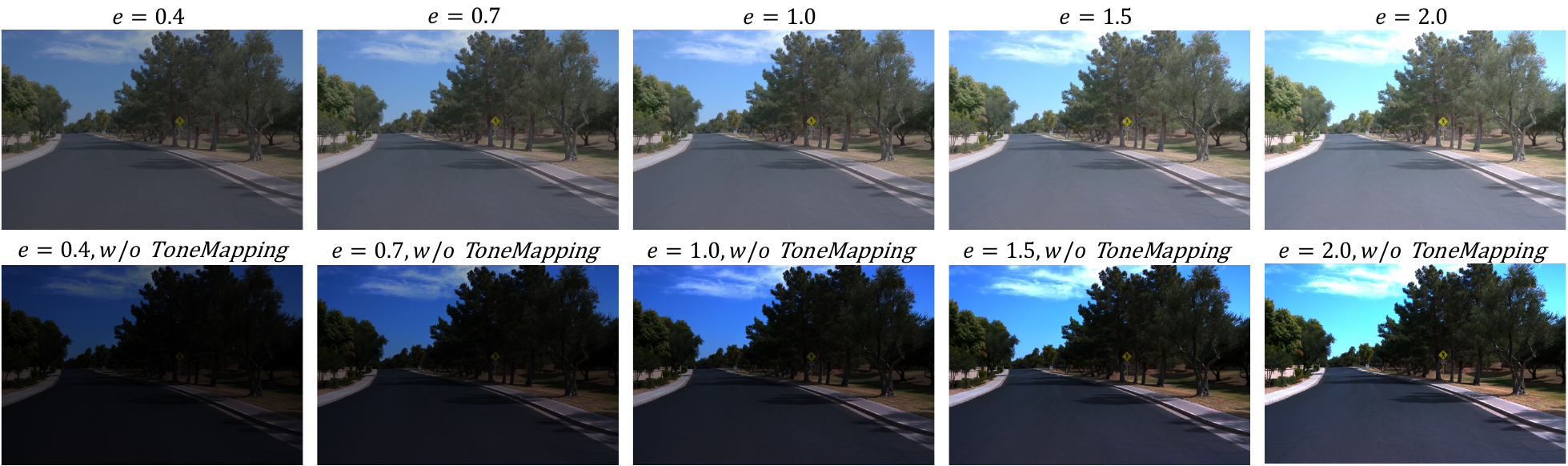}
\caption{Exposure scale and Tone-Mapping comparison on the Waymo Open Dataset. }
\label{ap_2}
\end{figure*}
Such controllability can be partially leveraged to emulate challenging conditions such as strong sunlight or adverse weather offering potential applications for evaluating autonomous driving models.

Importantly, our method supports arbitrary $e$ and $\gamma$ values \textbf{without compromising rendering speed}, ensuring scalability. 
Fig.\ref{ap_1} illustrates rendering results under different $\gamma$ values, while Fig.\ref{ap_2} compares renderings produced with varying $e$ values with and without tone-mapping.


\end{document}